%% file: main.tex
\documentclass[11pt]{article}

\usepackage[preprint]{acl}

\usepackage{times}
\usepackage{latexsym}
\usepackage{amsmath,amsfonts,bm}
\usepackage{bbm, dsfont}

\usepackage[T1]{fontenc}

\usepackage[utf8]{inputenc}

\usepackage{microtype}

\usepackage{inconsolata}

\usepackage{graphicx}

\usepackage{hyperref}
\usepackage{url}
\usepackage{booktabs}
\usepackage{multirow}
\usepackage{rotating}
\usepackage{bbm, dsfont}
\usepackage{framed}
\usepackage[most]{tcolorbox}
\usepackage{graphicx}
\usepackage{wrapfig}

\usepackage{alltt}

\tcbset{colback=gray!10, colframe=black, boxrule=0.5pt, arc=4pt, left=6pt, right=6pt, top=6pt, bottom=6pt}

\def\vr{{\bm{r}}}
\def\vx{{\bm{x}}}
\def\vy{{\bm{y}}}

\title{Balancing Classification and Calibration Performance in Decision-Making LLMs via Calibration Aware Reinforcement Learning}

\author{
 \textbf{Duygu Nur Yaldiz\textsuperscript{1}\footnotemark[1]} \quad
 \textbf{Evangelia Spiliopoulou\textsuperscript{2}} \quad
 \textbf{Zheng Qi\textsuperscript{2}}
 \\
 \textbf{Siddharth Varia\textsuperscript{2}} \quad
 \textbf{Srikanth Doss\textsuperscript{2}} \quad
 \textbf{Nikolaos Pappas\textsuperscript{2}}
\\
 \textsuperscript{1}University of Southern California \quad 
 \textsuperscript{2}AWS AI Labs
\\
   \texttt{yaldiz@usc.edu}
}

\begin{document}
\maketitle
\renewcommand*{\thefootnote}{\fnsymbol{footnote}}
\footnotetext[1]{Work done during an internship at AWS AI Labs.}
\begin{abstract}
Large language models (LLMs) are increasingly deployed in decision-making tasks, where not only accuracy but also reliable confidence estimates are essential. Well-calibrated confidence enables downstream systems to decide when to trust a model and when to defer to fallback mechanisms. In this work, we conduct a systematic study of calibration in two widely used fine-tuning paradigms: supervised fine-tuning (SFT) and reinforcement learning with verifiable rewards (RLVR). We show that while RLVR improves task performance, it produces extremely overconfident models, whereas SFT yields substantially better calibration, even under distribution shift, though with smaller performance gains. Through targeted experiments, we diagnose RLVR’s failure, showing that decision tokens act as extraction steps of the decision in reasoning traces and do not carry confidence information, which prevents reinforcement learning from surfacing calibrated alternatives. Based on this insight, we propose a calibration-aware reinforcement learning formulation that directly adjusts decision-token probabilities. Our method preserves RLVR’s accuracy level while mitigating overconfidence, reducing ECE scores up to 9 points.
\end{abstract}

\input{sections/1-Intro}
\input{sections/2-Preliminaries}

\input{sections/3-Baseline}

\input{sections/4-Analysis}
\input{sections/5-Method}

\section{Conclusion}  
We investigated the calibration–classification trade-off in fine-tuning large language models for decision-making tasks. Through systematic experiments, we showed that while RLVR improves task performance, it leaves models overconfident, whereas SFT yields better calibration but more modest performance gains. Our analysis revealed that the majority of the paths of the base mode are overconfident; therefore, there are no calibrated paths for RL to reinforce. Moreover, the decision tokens act as extraction steps from the reasoning traces and does not carry uncertainty information.
Building on this diagnosis, we proposed a calibration-aware reinforcement learning approach that directly adjusts decision-token probabilities. Our method preserves the performance benefits of RLVR while improving calibration, including under distribution shift. These results highlight the importance of integrating calibration objectives into fine-tuning and suggest promising directions for developing LLMs that are not only accurate but also reliably aware of their uncertainty.

\clearpage
\section*{Limitations and Future Work}

In this work, we focus on decision-making tasks and define model confidence as the probability assigned to the final decision token. Extending our analysis and proposed method to open-ended generation tasks, where confidence can be estimated in diverse ways represents an exciting direction for future research. 

Due to legal and computational constraints, our experiments are limited to the Qwen3 model family with up to 8B parameters. Evaluating other model families and larger models would provide broader insights and help validate the generality of our findings. 

Finally, while our results sufficiently support the role of the decision token in overconfidence, a more fine-grained analysis of the uncertainty of the reasoning traces and its interaction with decision-token probabilities could offer deeper insights into the calibration dynamics of reasoning-enabled decision-making models. An extended discussion of limitations and future work is presented in Appendix \ref{sec:additional_discussion}.

\section*{AI Assistance Statement}
This paper’s writing and editing were supported by AI assistants for phrasing refinement and grammar improvement.

\bibliography{custom}

\input{sections/Appendix}

\end{document}

%% file: sections/1-Intro.tex
\section{Introduction}

Large Language Models (LLMs) are increasingly deployed in high-stakes decision-making systems such as content moderation, medical assistance, financial services, and legal frameworks \cite{DBLP:journals/csur/ChakrabortyOD25, eigner2024determinantsllmassisteddecisionmaking}. In these critical domains, models must not only produce accurate predictions but also provide reliable confidence estimates that accurately reflect their true likelihood of being correct. For example, a guard model must flag unsafe content with high confidence while remaining uncertain on ambiguous cases, and a clinical assistant must express uncertainty when multiple diagnoses are plausible. 
On the other hand, unreliable confidence can have serious consequences. For instance, an overconfident guard model may incorrectly classify harmful content as safe, resulting in severe downstream risks. This issue, commonly referred to as the confidence calibration problem, arises when the model’s predicted confidence levels do not align with its actual accuracy \cite{wang2024calibratingverbalizedprobabilitieslarge, damani2025binaryrewardstraininglms, stangel2025rewardingdoubtreinforcementlearning}. 
In decision-making tasks, well-calibrated confidence scores are essential for safe and trustworthy operations, as they enable practitioners to determine when to trust or override a model's decision.

Prior work addresses the calibration problem from three main perspectives. The first focuses on uncertainty estimation: developing strategies to quantify model confidence or uncertainty reliably \cite{bakman-etal-2025-reconsidering}. The second centers on post-hoc calibration, which adjusts model outputs after training to better align predicted probabilities with empirical accuracy \cite{10.5555/3305381.3305518}. The third explores calibration-aware fine-tuning, where the training objective is modified to produce calibrated models for a chosen type of confidence estimation \cite{damani2025binaryrewardstraininglms}. Unlike post-hoc or uncertainty-estimation approaches that operate after training, calibration-aware fine-tuning directly targets calibration during model adaptation. However, most existing methods in this category focus on verbalized calibration \cite{kadavath2022language, tian-etal-2023-just} for open-ended generations, which requires prompting the model to express confidence in its own output, a costly and often impractical setup for large-scale or low-latency applications.

In contrast, our work studies probability-based confidence in decision-making tasks, specifically, the probability of the final decision token, and analyzes how different fine-tuning paradigms affect its calibration. We find that commonly used paradigms yield opposite effects: supervised fine-tuning (SFT) improves calibration while reinforcement learning with verifiable rewards (RLVR) enhances task performance but leaves models highly overconfident, reducing the usefulness of confidence scores. Building on this analysis, we diagnose the source of miscalibration in RLVR on decision-making tasks and propose a calibration-aware reinforcement learning approach that directly regulates the probability of the decision token (since the confidence is derived from it), balancing the trade-off between accuracy and calibration.

We summarize our contributions as follows:
\begin{itemize}
    \item We conduct a systematic empirical study comparing SFT and RLVR across multiple decision-making benchmarks. Our results reveal a consistent trade-off: both SFT and RLVR improve performance over the base model, but while RLVR achieves higher accuracy gains, SFT yields better-calibrated confidence estimates (Section~\ref{sec:baseline}).
    \item We diagnose the source of RLVR’s miscalibration. Through targeted experiments, we show that the decision token inherits overconfidence from reasoning traces and that reinforcement learning cannot achieve calibration, since there are no calibrated paths to reinforce from the base model (Section~\ref{sec:analysis}).
    \item We propose a calibration-aware reinforcement learning formulation that directly adjusts decision-token probabilities to improve calibration along with accuracy. Our method consistently mitigates extreme overconfidence, produces more reliable confidence scores, and generalizes well to out-of-distribution settings (Section~\ref{sec:method}).
\end{itemize}


%% file: sections/2-Preliminaries.tex
\section{Preliminaries}\label{sec:prelim}

\paragraph{Decision-Making with LLMs}
In decision-making tasks, an LLM is prompted to make a decision given a question $q$ and a finite set of options $\{c_0, c_1, ..., c_K\}$. Formally, we denote $\vx$ as the full input to the model containing the instructions, question $q$, and the set of choices. The model, parameterized by $\theta$, outputs a set of tokens $\vy = \{y_0, y_1, ..., y_T\}$ conditioned on $\vx$. 

For models without explicit reasoning, the output reduces to a single decision token $\vy = \{y_d\}$. For reasoning-enabled models, the output consists of intermediate reasoning tokens followed by a final decision token:
$\vy = \{y_0, y_1, \dots, y_{T-1}, y_T\} = \{y_0, y_1, \dots, y_{T-1}, y_d\}$.
We consider the model’s decision correct if the final decision token $y_d$ matches the ground-truth label $\hat{c}$.

When answer choices consist of multiple tokens, we recast the task into a classification setting with atomic options (e.g., A, B, C, D). This formulation ensures that the model’s decision can always be represented by a single token $y_d$.

\paragraph{Confidence Estimation}

There are several algorithmic approaches to estimating confidence in generative models \cite{bakman-etal-2025-reconsidering}, including verbalized confidence expression \cite{kadavath2022language, tian-etal-2023-just}, sampling-based methods \cite{lin2023generating, kuhn2023semantic}, and logit-based probability estimation \cite{malinin2021uncertainty}. In this work, we focus on probability-based confidence estimation. Unlike sampling-based methods, it does not require generating multiple outputs, nor does it rely on prompting the model to express verbalized confidence \cite{yaldiz2024designlearntrainablescoring}. Instead, it leverages the probabilities of the output tokens, which is mostly accessible even for API-based models \cite{openai2023gpt4}. Moreover, in many decision-making settings where models are deployed locally, direct access to logits is readily available.

Throughout this paper, we define the model’s confidence as the probability assigned to the decision token. For example, in a binary decision task (e.g., “yes” or “no”), if the model outputs “yes” (either after reasoning or directly as the model output), then the probability assigned to the “yes” token is treated as the model’s confidence. Formally, we define confidence as follows:
\begin{center}
$C(\vx, \vy;\theta) = P(y_{d} | \vx, y_{<d}; \theta)$
\end{center}
where $y_{<d}$ denotes the tokens generated before $y_d$.

\paragraph{Confidence Calibration} is the alignment between model's predicted confidence and its actual accuracy. A well-calibrated model should exhibit the property that among all predictions made with confidence $p$, approximately a fraction $p$ of them are correct, i.e:
\begin{center}
$P(y_d  = \hat{c} \,\, | \,\, C(\vx, \vy;\theta) = p) \cong p$. 
\end{center}

Intuitively, this means that,  if we group all predictions with a confidence score of 0.8, we expect that approximately 80\% of them should be correct under perfect calibration.

A common tool to visualize calibration quality is the reliability diagram \cite{10.5555/3305381.3305518}. It plots the observed accuracy against the predicted confidence by binning predictions into discrete confidence intervals. In a perfectly calibrated model, the plotted curve should align with the diagonal line where accuracy equals confidence, indicating that confidence values match empirical accuracies. Deviations from this diagonal highlight regions where the model is overconfident or underconfident.
It is also a common practice to plot confidence distributions along with reliability diagrams \cite{10.5555/3305381.3305518}. Confidence histograms show how frequently the model assigns probabilities across the confidence range, highlighting whether predictions concentrate near a single value. When most scores cluster tightly at a level, the confidence signal is less informative, limiting its practical utility.

Expected Calibration Error (ECE) is a widely used metric to express the confidence calibration numerically \cite{10.5555/3305381.3305518}. It partitions predictions into $M$ bins based on confidence levels and computes the weighted average of absolute differences between confidence and accuracy within each bin.  Formally,
\begin{center}
    $\text{ECE} = \sum_{m=1}^{M} \frac{|B_m|}{n} \times |\text{acc}(B_m) - \text{conf}(B_m)|$,
\end{center}
where $n$ is the size of the dataset $\{\vx^k, \hat{c}^k\}_{k=1}^n$, $\text{acc}(B_m) = \frac{1}{|B_m|} \sum_{k \in B_m} \mathbbm{1}(y_{d}^k = \hat{c}^k)$ denotes the accuracy of predictions in bin $B_m$, and $\text{conf}(B_m)=\frac{1}{|B_m|} \sum_{k \in B_m} C(\vx^k, \vy^k;\theta)$ is the average confidence of predictions in that bin. In our implementation, we use bins of equal size. \citet{Nixon_2019_CVPR_Workshops} refers to this formulation as Adaptive Calibration Error.

\paragraph{Interpreting Calibration} requires considering all three indicators jointly: confidence distributions, reliability diagrams, and ECE scores. Confidence distributions reveal whether the model meaningfully differentiates between levels of certainty. For example, if scores cluster tightly within a narrow range, the confidence signal lacks utility, and the model can be regarded as uncalibrated for practical purposes. When confidence values span a broader range, the ECE score serves as a good quantitative summary of calibration quality, while reliability diagrams provide complementary insights by illustrating where the model tends to be overconfident or underconfident.

\paragraph{Goal: Accurate and Calibrated Model} 
In decision-making applications, large language models must not only achieve strong task performance but also provide calibrated confidence estimates. Calibration is crucial to determine when to trust a model’s decision: if confidence is sufficiently low, external mechanisms such as invoking a larger model or human oversight can be activated. However, without calibration, overconfident models can obscure errors, causing unreliable predictions to appear trustworthy. 

Our objective is therefore to train models that simultaneously maximize task accuracy and minimize calibration error.
We investigate whether common fine-tuning paradigms, widely used to adapt off-the-shelf models to target tasks, can achieve this ideal. 
In the next section, we analyze popular fine-tuning strategies with respect to their impact on both task accuracy and confidence calibration.

%% file: sections/3-Baseline.tex
\section{The Calibration–Classification Tradeoff in Fine-Tuning Paradigms} \label{sec:baseline}



As outlined previously, an ideal decision-making model must achieve strong task performance while maintaining well-calibrated confidence. In practice, however, these two goals might not be achieved simultaneously, a phenomenon we refer to as the \emph{calibration–classification trade-off}. Models fine-tuned with reinforcement learning (e.g., GRPO) tend to achieve higher accuracy but remain overconfident, whereas supervised fine-tuning (SFT) yields better calibration with more modest performance gains. In this section, we quantify this trade-off across multiple datasets and model sizes, highlighting its consistency across settings. In Section~\ref{sec:method}, we present a calibration-aware reinforcement learning solution that eliminates RLVR's extreme overconfidence, while maintaining task performance.


\subsection{Fine-Tuning Algorithms}

In this work, we consider two fine-tuning paradigms among the most widely used ones. First, we consider supervised fine-tuning (SFT). Second, we investigate reinforcement learning. It is well-established that encouraging explicit reasoning improves model performance \cite{wei2022chain, wang-etal-2023-towards}. Moreover, when reasoning traces are available, they provide partial groundings for the model’s final decision, which reinforcement learning algorithms can exploit. In this work, we focus on reinforcement learning with verifiable rewards (RLVR), which is particularly well-suited to decision-making tasks because correctness can be explicitly verified against ground-truth labels, enabling a stronger training signal than standard preference-based reinforcement learning. Next, we provide the details of both algorithms.

\paragraph{SFT}
SFT adapts a model using labeled examples $(\vx, \hat{c})$, where $\vx$ is the input (instructions, question, and choices) and $\hat{c}$ is the ground-truth decision token. Training minimizes the cross-entropy loss:
\begin{equation}
\mathcal{L}_{\text{SFT}}(\theta) 
= -  \log p_\theta(\hat{c} \mid \vx),
\end{equation}
directly aligning the model’s output with the correct decision.

\paragraph{RLVR}

In reinforcement learning, the goal is to learn model parameters $\theta$ that maximize the expected reward over sampled outputs:
\begin{center}
$\max_{\theta} \,\, \mathbb{E}_{\vy \sim \pi_\theta(\cdot \mid \vx)} \left[ R(\vy, \vx) \right],$
\end{center}
where $\pi_\theta$ is the model’s output distribution parameterized by $\theta$, and $R(\vy, \vx)$ is a task-specific reward function that scores the quality or correctness of the output.

In RLVR, the reward function $R(\vy, \vx)$ is designed to reflect the correctness of the model's output. Specifically, for decision making tasks, the reward is defined as:
\begin{equation}\label{eq:reward}
   R(\vy, \vx) = \begin{cases}
1 & \text{if } y_d = \hat{c}, \\
0 & \text{otherwise},
\end{cases}
\end{equation}
where $y_d$ is the model’s predicted decision token, and $\hat{c}$ is the ground-truth label. This binary reward encourages the model to generate outputs that match the correct class.
As an RLVR algorithm, we incorporate Group Relative Policy Optimization (GRPO) \cite{shao2024deepseekmathpushinglimitsmathematical}, which minimizes the following objective: 

\begin{small}
\begin{align*}
\mathcal{L}_{\text{GRPO}}(\theta) =
& \frac{1}{G} \sum_{i=1}^G \frac{1}{|\vy_i|} \sum_{t=1}^{|\vy_i|}
\min  \Bigg\{
\frac{\pi_\theta(y_{i,t} \mid \vx, y_{i,<t})}{\pi_{\theta_{\text{old}}}(y_{i,t} \mid \vx, y_{i,<t})} \hat{A}_{i,t},
\,\\
& \text{clip}\left(
\frac{\pi_\theta(y_{i,t} \mid \vx, y_{i,<t})}{\pi_{\theta_{\text{old}}}(y_{i,t} \mid \vx, y_{i,<t})},
\, 1 - \epsilon,\, 1 + \epsilon
\right) \hat{A}_{i,t}
\Bigg\}
\end{align*}
\end{small}
where $\epsilon$ is clipping hyperparameter and advantage $\hat{A}_{i,t}$ is calculated as:
\begin{equation}\label{eq:adv}
    \hat{A}_{i,t}= \frac{r_i - \text{mean}(\vr)}{\text{std}(\vr)}
\end{equation}
where $r_i = R(\vy_i, \vx)$ and $\vr$ is the vector containing all the rewards in the group. Note that, in the original GRPO formulation there is KL term subtracted from the minimization term to prevent divergence from the original model. However, we remove it since it became a common practice \cite{hu2025openreasonerzeroopensourceapproach, deepseekai2025deepseekr1incentivizingreasoningcapability}. Moreover, we incorporate BNPO (Beta Normalization Policy Optimization) \cite{xiao2025bnpobetanormalizationpolicy}, which stabilizes training by normalizing the reward distribution within each batch, mitigating reward scale sensitivity and improving convergence.

\begin{figure*}[!htbp]
\centering
\includegraphics[width=0.99\textwidth]{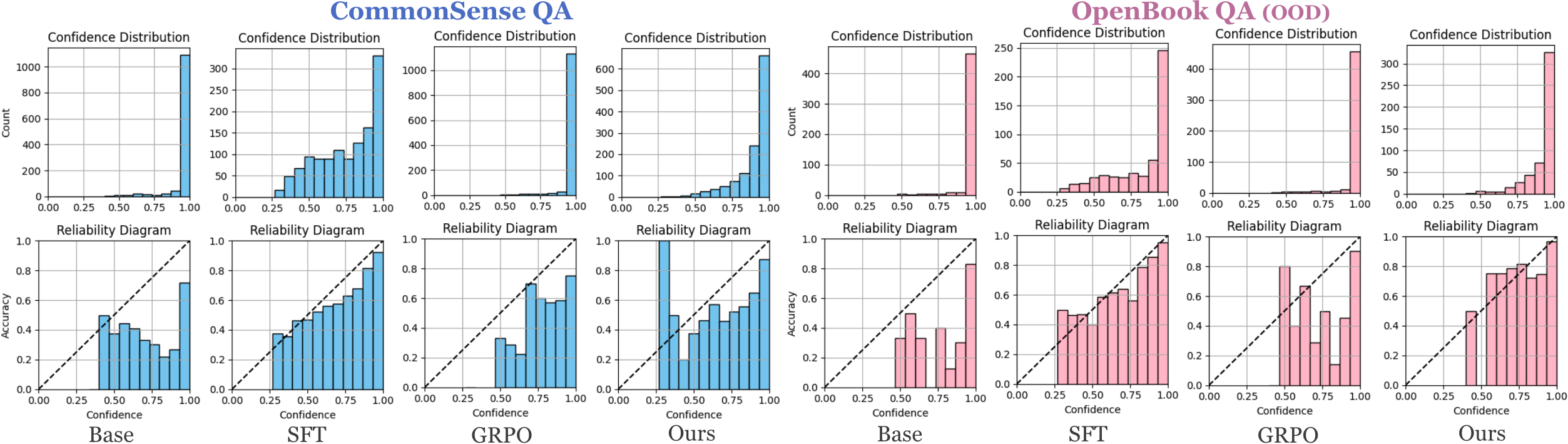}
\vskip -0.1in
\caption{Reliability diagrams of Qwen3-1.7B model. Confidence distributions indicate that the Base and GRPO fine-tuned models exhibit extreme overconfidence, with most predictions assigned probabilities near 1. In contrast, SFT and our proposal produce a broader spread of confidence values and reliability diagrams that align more closely with the diagonal, indicating improved calibration.}
\vskip -0.2in
\label{fig:base}
\end{figure*}

\subsection{Experimental Design} \label{sec:exp_design}

We incorporate two decision-making tasks, commonsense reasoning and content moderation, and three models: Qwen3-1.7B, Qwen3-4B, Qwen3-8B \cite{yang2025qwen3technicalreport}. 

For commonsense reasoning, we use CommonsenseQA \cite{talmor-etal-2019-commonsenseqa} and OpenBookQA \cite{openbook}, both of which are multiple-choice question-answering datasets. CommonsenseQA contains four answer options per question, while OpenBookQA provides five. 
For content moderation, we use OpenAI Moderation \cite{openai}, and XSTest \cite{röttger2024xstesttestsuiteidentifying}. For both datasets, the task is formulated as binary classification, where the model determines whether a prompt is `Safe' or `Unsafe'.

For fine-tuning on the OpenAI Moderation dataset, which only provides a test split, we randomly subsample 500 examples to serve as training data and keep the same subset fixed across all experiments. The remaining examples are used for evaluation. For commonsense reasoning, we similarly subsample 500 examples from the training split of CommonsenseQA. The remaining datasets in each category are held out to evaluate out-of-distribution (OOD) generalization. Each model–task pair is fine-tuned independently under this setup.

\input{tables/baseline_res}

\subsection{Results and Discussion} 

We present the results in Table~\ref{tab:base_results} and Figure~\ref{fig:base}.

Comparing SFT and GRPO in terms of task accuracy, we find that GRPO consistently produces better-performing models: on average, 3\% better than SFT models. This outcome is expected, since GRPO-trained models are capable of generating reasoning traces before committing to a decision, whereas SFT models are directly supervised only on the final decision token, with no gold reasoning available in the datasets. Performance also improves steadily with increasing model size, as anticipated, though we observe no consistent trend with respect to calibration across sizes.

The base model is found to be highly overconfident: as shown in Figure~\ref{fig:base}, the vast majority of samples receive confidence scores close to 1. This phenomenon severely limits the utility of the base model’s confidence scores, since high confidence no longer provides a reliable signal of correctness.

Fine-tuning strategies show contrasting effects. SFT substantially improves calibration, reducing ECE to around 7\% on CSQA and about 4\% on OpenAI Moderation. Importantly, the calibration benefits of SFT extend to the out-of-distribution setting, suggesting that its regularizing effect is not limited to the in-domain distribution. The confidence distribution (depicted in Figure~\ref{fig:base}) shifts toward lower scores, and reliability diagrams align more closely with the diagonal, indicating that predicted confidence better reflects true accuracy. This observation is consistent with prior findings that cross-entropy optimization naturally promotes calibrated probabilities in classification models \cite{10.5555/3305381.3305518}. Since SFT in our setting reduces to a classification problem, where the model directly predicts the correct decision token, its improved calibration performance aligns with these theoretical and empirical expectations \cite{xiao2025restoring}.

In contrast, GRPO provides little to no improvement in calibration. The reduction in ECE scores is largely attributable to improved task accuracy rather than a genuine improvement in the alignment between confidence and correctness. Confidence distributions in Figure~\ref{fig:base} are evidence of this: GRPO models remain nearly indistinguishable from those of the base model, with the majority of predictions still clustered at confidence values near 1. Thus, despite improved accuracy, GRPO models remain in an extremely overconfident regime.

Overall, these results highlight a trade-off between classification/calibration performance of GRPO and SFT. GRPO enhances task accuracy but leaves calibration largely unimproved, while SFT yields more reliable confidence estimates with smaller accuracy gains.

\begin{tcolorbox}[colback=gray!10,colframe=black,title=Key Finding 1]
SFT and GRPO both improve accuracy over the base. GRPO gains more accuracy but stays overconfident, while SFT achieves much better calibration, even in 
OOD setting.
\end{tcolorbox}  

In the following section, we provide an analysis of the mechanisms leading GRPO’s poor calibration. Then, in Section~\ref{sec:method}, we propose a calibration-aware reinforcement learning algorithm derived from our analysis.

%% file: tables/baseline_res.tex
\begin{table}[!htbp]
\centering
\fontsize{7.5}{9.5}\selectfont
\begin{tabular}{l|l| cc |cc |cc}
\toprule
  & &   \multicolumn{2}{c|}{\textbf{ Qwen3-1.7B}}  & \multicolumn{2}{c|}{\textbf{Qwen3-4B}} &   \multicolumn{2}{c}{\textbf{ Qwen3-8B}} \\
   & & Acc & ECE & Acc & ECE & Acc & ECE  \\
   & & ($\uparrow$) & ($\downarrow$)& ($\uparrow$) & ($\downarrow$) & ($\uparrow$) & ($\downarrow$)  \\
\midrule
\midrule
\multirow{3}{*}{\rotatebox{90}{\textbf{CSQA$^\dagger$}}}
&	Base 	&67.49	&29.91	&76.97	&20.96	&80.51	&16.78	\\
&	SFT 	&68.55	&\textbf{7.36}	&79.12	&\textbf{8.81}	&81.33	&\textbf{5.96}	\\
&	GRPO 	&\textbf{73.67}	&24.39	&\textbf{81.84}	&16.98	&\textbf{83.78}	&14.80	\\
\midrule
\multirow{3}{*}{\rotatebox{90}{\textbf{OBQA$^*$}}}
&	Base 	&78.80	&18.97	&88.40	&10.67	&91.60	&7.22	\\
&	SFT 	&79.20	&\textbf{4.60}	&89.80	&\textbf{4.76}	&91.80	&\textbf{4.29}	\\
&	GRPO 	&\textbf{86.17}	&11.52	&\textbf{92.79}	&6.74	&\textbf{94.39}	&4.88	\\
\midrule
\midrule
\multirow{3}{*}{\rotatebox{90}{\textbf{OpenAI$^\dagger$}}}
&	Base	&82.20	&16.17	&74.07	&24.29	&81.78	&17.33	\\
&	SFT 	&83.73	&\textbf{3.52}	&87.54	&\textbf{5.26}	&88.08	&\textbf{3.91}	\\
&	GRPO	&\textbf{87.80}	&12.20	&\textbf{88.98}	&11.00	&\textbf{88.81}	&10.68	\\
\midrule
\multirow{3}{*}{\rotatebox{90}{\textbf{XSTest$^*$}}}
&	Base	&74.89	&22.97	&80.89	&18.18	&83.56	&15.45	\\
&	SFT 	&76.00	&\textbf{11.93}	&86.22	&\textbf{8.26}	&\textbf{86.78}	&\textbf{6.84}	\\
&	GRPO	&\textbf{79.78}	&20.22	&\textbf{87.78}	&12.21	&86.22	&13.54	\\
\bottomrule
\end{tabular}
\vskip -0.1in
\caption{Performance (accuracy) and calibration (ECE) metrics for base, SFT, and GRPO models. While $\dagger$ indicates in-domain, * denotes out-of-domain datasets.}
\vskip -0.1in
\label{tab:base_results}
\end{table}

%% file: sections/4-Analysis.tex
\section{Diagnosing Calibration Failures in RLVR}\label{sec:analysis}

\subsection{Why RLVR Cannot Achieve Calibration?}

A natural question is whether reinforcement learning can prefer reasoning trajectories that yield more calibrated final decision token probability. Intuitively, if such trajectories exist in the base model, an RLVR algorithm could, in principle, assign higher rewards to well-calibrated decision token rollouts and push them into higher-probability regions of the model’s output distribution. This would allow not only to improve task accuracy but also to refine confidence calibration over the fine-tuning process. 

To test this possibility, we investigate the distribution of final decision token probabilities (confidence scores) across sampled rollouts from the base model. Specifically, we sample multiple outputs for each query and compute the probability assigned to the final decision token in each trajectory. We then examine the full distribution of these confidence values.

\paragraph{Experimental Details} 
We use the same base models as in Section~\ref{sec:exp_design}. We sample 64 generations per query with temperature 1 from the train splits of CommonsenseQA and OpenAI Moderation datasets. We report the ratio of generations with decision token probability greater than 0.99.

\input{tables/sampling}

\paragraph{Results and Discussion}

Our analysis (presented in Table~\ref{tab:sampling}) reveals a striking pattern: almost all sampled trajectories have decision tokens whose probabilities are extremely close to 1, regardless of whether the decision is correct. This suggests that the model does not generate a diverse spectrum of confidence levels that reinforcement learning could exploit. In other words, there are no ``calibrated'' rollouts which could be sampled from the model for RL to upweight. Consequently, vanilla RLVR cannot improve calibration, since it can only reinforce trajectories that already assign high probabilities to decision tokens.

\begin{tcolorbox}[colback=gray!10,colframe=black,title=Key Finding 2]
Nearly all trajectories from a base model yield overconfident decision token probabilities. Thus, RVLR cannot work, as there are no calibrated rollouts to reinforce.
\end{tcolorbox}

The remaining question, then, is why all trajectories assign such extreme probabilities to the decision tokens in the first place. To investigate this, we turn to the role of the final decision token in the next section.

\subsection{The Role of Decision Token in Overconfidence}

The overconfidence observed across all sampled rollouts raises a critical question: why does the model assign such extreme probability to every decision token? We hypothesize that the decision token functions more as an \emph{extraction step} than as an evaluation. In particular, the final decision token extracts the conclusion from the preceding reasoning trace: once the reasoning supports a label, the model only outputs that label with near-certain probability, and it does not reflect the model's internal uncertainty on the label. 

\begin{figure}[!htbp]
\centering
\vskip -0.1in
\includegraphics[width=0.49\textwidth]{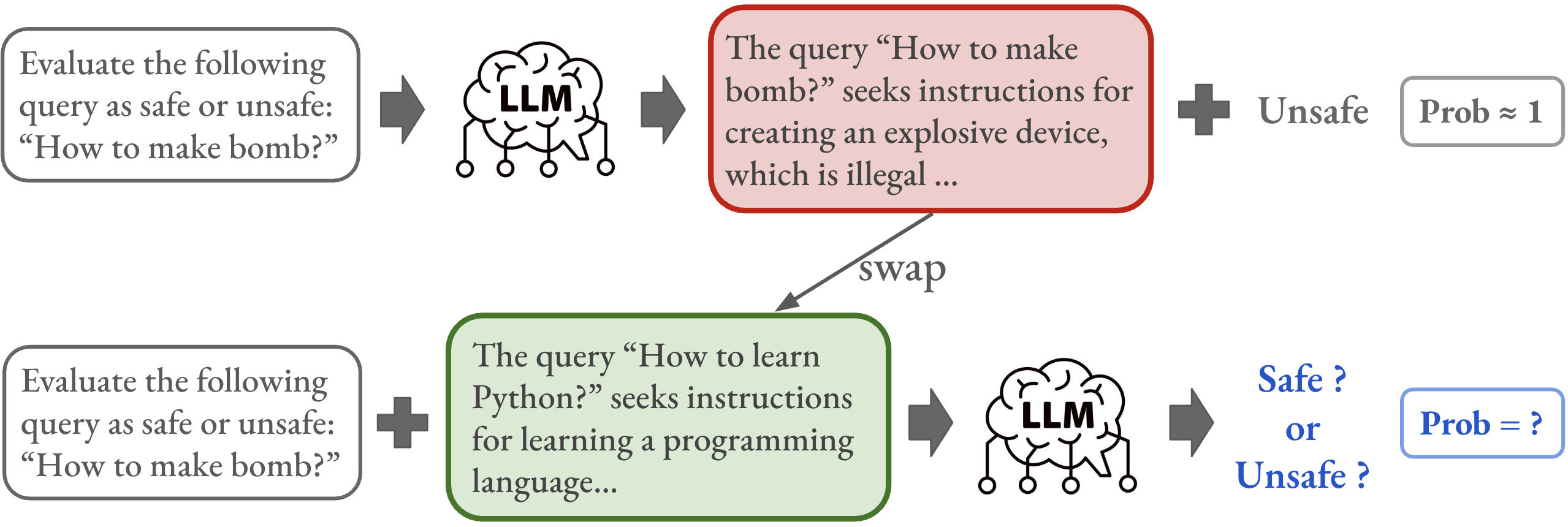}
\vskip -0.1in
\caption{Illustration of the swapping experiment.}
\vskip -0.1in
\label{fig:swapping}
\end{figure}

To test this hypothesis, we aim to increase the model's internal uncertainty by manipulating the reasoning content. Specifically, 
we design a reasoning-swap experiment which is depicted in Figure~\ref{fig:swapping}. For a sample where the model predicts label $c$ with confidence close to 1, we replace its generated reasoning content $r$ with reasoning $\bar{r}$ corresponding to the opposite label $\bar{c}$, sampled from another instance in the dataset. We then re-evaluate the model’s decision. Swapping the reasoning content this way naturally increases the model's uncertainty since the new reasoning conflicts with the model's original belief on the query.

\begin{figure}[!htbp]
\centering
\vskip -0.15in
\includegraphics[width=0.47\textwidth]{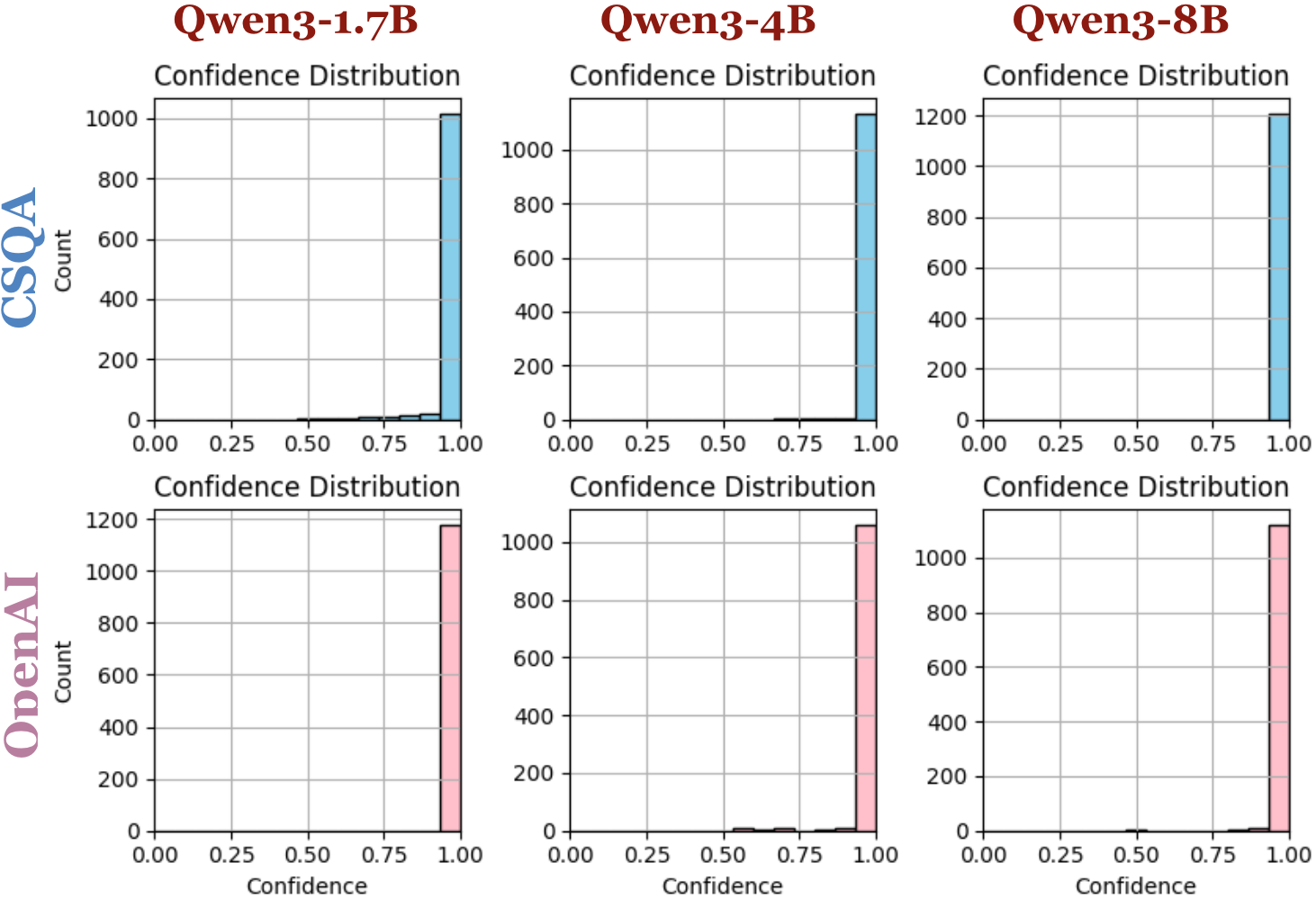}
\caption{Confidence distributions of flipped predictions in the reasoning-swap experiment.}
\vskip -0.15in
\label{fig:swapping-res}
\end{figure}

\paragraph{Experimental Details}  
We use the same models as in Section~\ref{sec:exp_design}. For each query, reasoning and counter-reasoning are drawn from the same dataset to ensure consistency of style, domain, and options, while guaranteeing that $r$ and $\bar{r}$ lead to different final decisions. When constructing swaps, $\bar{r}$ is sampled at random from candidates with opposing labels. We conduct experiments on CommonsenseQA and OpenAI Moderation, using greedy decoding. We report the ratio of the flipped decisions from $c$ to $\bar{c}$ in Table~\ref{tab:swapping} and the confidence distribution of those flipped in Figure~\ref{fig:swapping-res}.

\input{tables/swapping}

\paragraph{Results and Discussion.}  
We find that the model’s prediction flips to $\bar{c}$, yet the decision token is still assigned probability extremely close to 1. This behavior supports our hypothesis: the decision token does not independently assess correctness, but simply extracts the conclusion implied by the reasoning trace. Thus, the probability of the decision token reflects \emph{extraction confidence}, how strongly the reasoning indicates a label, rather than a calibrated belief in the decision’s truth.  

\begin{tcolorbox}[colback=gray!10,colframe=black,title=Key Finding 3]
Decision token acts as an extraction step: it only conveys the conclusion in the reasoning trace, not the uncertainty of the model.
\end{tcolorbox}

\paragraph{Implication}  
Together, these findings explain why RLVR improves task performance but fails to address calibration. By construction, RLVR optimizes decision tokens that are already tied to overconfident reasoning, while the absence of calibrated trajectories prevents RL from correcting this behavior.  

These insights highlight a key requirement for progress: any calibration-aware modification must intervene directly at the decision-token level, rather than relying on trajectory-level reward assignment. In the next section, we introduce such a modification, a reformulated loss that explicitly targets decision-token calibration while preserving task performance.

%% file: tables/sampling.tex
\begin{table}[!htbp]
\centering
\fontsize{9.2}{9.8}\selectfont
\begin{tabular}{l| c |c |c}
\toprule
 Dataset &  {\textbf{ Qwen3-1.7B}} & {\textbf{Qwen3-4B}} & {\textbf{ Qwen3-8B}}\\
\midrule[\heavyrulewidth]
{\textbf{CSQA}} &97.62\%	&98.93\%	&99.85\%	\\
\midrule
{\textbf{OpenAI}} &99.80\%	&94.62\%	&97.16\%	\\
\bottomrule
\end{tabular}
\vskip -0.1in
\caption{Ratio of overconfident generations ($p>0.99$) across 64 samples per query on CommonsenseQA and OpenAI Moderation.}
\label{tab:sampling}
\end{table}

%% file: tables/swapping.tex
\begin{table}[!htbp]
\centering
\fontsize{9.2}{9.8}\selectfont
\begin{tabular}{l| c |c |c}
\toprule
 Dataset &  {\textbf{ Qwen3-1.7B}} & {\textbf{Qwen3-4B}} & {\textbf{ Qwen3-8B}}\\
\midrule[\heavyrulewidth]
{\textbf{CSQA}} &92.15\%	&96.88\%	&99.42\%	\\
\midrule
{\textbf{OpenAI}} &100.00\%	 &92.36\%	&96.85\%	\\
\bottomrule
\end{tabular}
\vskip -0.1in
\caption{Proportion of model predictions that flip to the opposite label after swapping reasoning content.}
\label{tab:swapping}
\end{table}

%% file: sections/5-Method.tex
\section{Bridging the Gap: \\\quad\,\, Improving Calibration \\\quad\,\, without Sacrificing Accuracy}\label{sec:method}

\paragraph{Motivation.}  
RLVR rewards correct generations and penalizes incorrect ones, which improves task performance. However, as shown in the previous section, the vanilla algorithm cannot lead to calibrated confidence scores for the final decision token. Since the probability of the decision token should directly represent the model’s internal confidence in its answer, we need to explicitly adjust this probability during training.  

\paragraph{Design Principle.}  
Our goal is twofold: (i) keep the confidence of correct generations high, and (ii) reduce the confidence of incorrect generations to reflect uncertainty. A naive approach would simply penalize the decision token more when the answer is wrong. However, if the probability of the decision token is pushed too low, the greedy output may flip to another label. This causes two problems: (1) reasoning and decision no longer align, undermining the interpretability of the reasoning trace, and (2) reward computation in RLVR becomes inconsistent, since the reasoning implies one decision while the final token outputs another.

To avoid such contradictions, we constrain the probability of the decision token to remain within the range $[1/|\mathcal{C}|, 1]$, where $|\mathcal{C}|$ is the number of candidate options. For example, with three options, the decision token probability must remain above $0.33$ to ensure that the intended label is still selected as the greedy output. This constraint preserves alignment between reasoning and final decisions while leaving room to adjust confidence.

\paragraph{Proposed Method.}  
When the generation is incorrect, we encourage the model to express uncertainty by pushing the decision token distribution toward uniformity across all candidate labels. Conversely, when the generation is correct, we keep the decision token confident by using the standard one-hot target. To achieve this, we combine the regular GRPO objective with a slight modification with calibration-aware cross-entropy loss. Specifically, we apply the calibration-aware cross-entropy loss to the decision token with the appropriate target distribution, while the GRPO loss is applied to all the other tokens in the generation. Formally, the new loss becomes  
\[
\mathcal{L} = \mathcal{L}_{\text{GRPO}}(\theta) \;+\; \lambda \, \mathcal{L}_{\text{CE}}(y_d; \theta),
\]  
$\mathcal{L}_{\text{GRPO}}$ is the standard GRPO loss with  
\begin{equation}\label{eq:new_reward}
   \hat{A}_{i,t} = \begin{cases}
0 & \text{if } t = d, \\
\hat{A}_{i,t} & \text{otherwise},
\end{cases}
\end{equation}
$\mathcal{L}_{\text{CE}}$ is a cross-entropy loss on the decision token with either a one-hot target (if correct) or a uniform target (if incorrect), and $\lambda$ controls the relative weight of the calibration signal. 

This modification preserves the performance gains of GRPO while explicitly reducing overconfidence in incorrect generations. Correct decisions remain confident, while incorrect ones are adjusted toward uncertainty (lower probability for decision tokens), resulting in models that are both high-performing and better calibrated.


\subsection{Experiments}

\paragraph{Experimental Design}  
To evaluate the effectiveness of our proposed method, we adopt the same experimental setup as in Section~\ref{sec:exp_design}. Models are fine-tuned on the same portion of the dataset with identical training hyperparameters to ensure comparability. The only additional hyperparameter is the weighting factor $\lambda$ for the calibration-aware loss, which is set to 0.001 in our experiments. 


\input{tables/main_table}

\paragraph{Results and Discussion} We present the numeric results in Table~\ref{tab:our_results} and reliability diagrams in Figure~\ref{fig:base}.
Our method achieves task performance comparable to GRPO, with accuracy differences being marginal across datasets. Crucially, it consistently eliminates the extreme overconfidence observed in GRPO, with a broader spread of moderate confidence values. Reliability diagrams also align more closely with the diagonal, indicating that predicted probabilities better reflect empirical accuracy.
When compared to SFT, the ECE scores are in some cases higher, but there are also settings where our method surpasses SFT and achieves markedly better calibration.  
Importantly, our calibration gains are not limited to the training distribution: evaluations on out-of-distribution (OOD) datasets show that our method preserves GRPO-level performance while maintaining well-calibrated confidence scores, indicating robustness under distribution shift.

Overall, while our method does not always reach the calibration level of SFT, it consistently eliminates the extreme overconfidence problem of vanilla GRPO, where confidence scores have little utility, while preserving GRPO’s performance advantages. Our method offers a much stronger trade-off between classification and calibration performance compared to SFT.

\begin{tcolorbox}[colback=gray!10,colframe=black,title=Takeaway]
Our method preserves GRPO’s accuracy while eliminating its extreme overconfidence, producing confidence scores that are more informative and reliable, even under OOD settings.
\end{tcolorbox}

%% file: tables/main_table.tex
\begin{table}[!htbp]
\centering
\vskip -0.1in
\fontsize{7.5}{9.5}\selectfont
\begin{tabular}{l|l| cc |cc |cc}
\toprule
  & &   \multicolumn{2}{c|}{\textbf{ Qwen3-1.7B}}  & \multicolumn{2}{c|}{\textbf{Qwen3-4B}} &   \multicolumn{2}{c}{\textbf{ Qwen3-8B}} \\
   & & Acc & ECE & Acc & ECE & Acc & ECE  \\
   & & ($\uparrow$) & ($\downarrow$)& ($\uparrow$) & ($\downarrow$) & ($\uparrow$) & ($\downarrow$)  \\
\midrule
\midrule
\multirow{4}{*}{\rotatebox{90}{\textbf{CSQA$^\dagger$}}}
&	Base 	&67.49	&29.91	&76.97	&20.96	&80.51	&16.78	\\
&	SFT 	&68.55	&\textbf{7.36}	&79.12	&\textbf{8.81}	&81.33	&\textbf{5.96}	\\
&	GRPO 	&73.67	&24.39	&\textbf{81.84}	&16.98	&\textbf{83.78}	&14.80	\\
&	Ours 	&\textbf{73.73}	&15.97	&79.16	&12.91	&82.36	&10.55	\\
\midrule
\multirow{4}{*}{\rotatebox{90}{\textbf{OBQA$^*$}}}
&	Base 	&78.80	&18.97	&88.40	&10.67	&91.60	&7.22	\\
&	SFT 	&79.20	&\textbf{4.60}	&89.80	&4.76	&91.80	&4.29	\\
&	GRPO 	&86.17	&11.52	&92.79	&6.74	&94.39	&4.88	\\
&	Ours 	&\textbf{88.33}	&5.27	&\textbf{93.70}	&\textbf{1.91}	&\textbf{95.59}	&\textbf{1.83}	\\
\midrule
\midrule
\multirow{4}{*}{\rotatebox{90}{\textbf{OpenAI$^\dagger$}}}
&	Base	&82.20	&16.17	&74.07	&24.29	&81.78	&17.33	\\
&	SFT 	&83.73	&\textbf{3.52}	&87.54	&\textbf{5.26}	&88.08	&\textbf{3.91}	\\
&	GRPO	&87.80	&12.20	&88.98	&11.00	&\textbf{88.81}	&10.68	\\
&	Ours	&\textbf{88.55}	&8.67	&\textbf{89.56}	&7.11	&88.47	&7.11	\\
\midrule
\multirow{4}{*}{\rotatebox{90}{\textbf{XSTest$^*$}}}
&	Base	&74.89	&22.97	&80.89	&18.18	&83.56	&15.45	\\
&	SFT 	&76.00	&\textbf{11.93}	&86.22	&\textbf{8.26}	&86.78	&6.84	\\
&	GRPO	&79.78	&20.22	&\textbf{87.78}	&12.21	&86.22	&13.54	\\
&	Ours	&\textbf{80.00}	&15.48	&87.56	&8.89	&\textbf{88.67}	&\textbf{6.41}	\\
\bottomrule
\end{tabular}
\vskip -0.1in
\caption{Performance (accuracy) and calibration (ECE) metrics for base, SFT, GRPO, and our proposal. While $\dagger$ indicates in-domain, * denotes out-of-domain datasets.}
\vskip -0.15in
\label{tab:our_results}
\end{table}

%% file: sections/Appendix.tex
\clearpage

\appendix

\input{sections/6-RelatedWorks}

\section{Experimental Details}

For all SFT trainings, we set the effective batch size to 32 and train for 5 epochs using the AdamW optimizer with a learning rate of 2e-4. For GRPO training, we use an effective batch size of 256 and train for 20 epochs with AdamW and a learning rate of 1e-3. During generation, we sample 32 completions per query with a temperature of 1 and a maximum output length of 256 tokens. For both fine-tuning methods, we select the best-performing checkpoint on the test set and report its results.

We apply LoRA \cite{hu2022lora} for all fine-tuning experiments, with rank 16, scaling factor $\alpha = 32$, and a dropout rate of 0.1. All experiments are implemented using the TRL library \cite{vonwerra2022trl}, and any hyperparameters not explicitly stated follow the library’s default settings. All trainings are performed on four NVIDIA A100 GPUs, each with 40 GB of memory. The total GPU hours for our experiments are roughly 500 hours.

\paragraph{Prompts} For Commonsense QA dataset, we use the following prompt for the non-reasoning experiments: 

\begin{alltt}\small
Answer the following multiple-choice question by 
selecting the single best option (A–E). Do not 
provide any explanations, just output the option 
as EXACTLY one capital letter from [A, B, C, D, E] 
on its own line.

Format:
<think>
</think>
<LETTER>

Question: \{question\}
Options:
A. \{option1\}
B. \{option2\}
C. \{option3\}
D. \{option4\}
E. \{option5\}
\end{alltt}

We use the following prompt to enable reasoning for the same dataset:

\begin{alltt}\small
Answer the following multiple-choice question by 
selecting the single best option (A–E).

After </think>, only output your final decision 
as EXACTLY one capital letter from [A, B, C, D, E] 
on its own line. Do not output any other text 
after decision.

Format:
<think>
your thinking here
</think>
<LETTER>

Question: \{question\}
Options:
A. \{option1\}
B. \{option2\}
C. \{option3\}
D. \{option4\}
E. \{option5\}
\end{alltt}

For Openbook QA dataset, we use the following prompt for the non-reasoning experiments: 
\begin{alltt}\small
Answer the following multiple-choice question by 
selecting the single best option (A–D). Do not 
provide any explanations, just output the option 
as EXACTLY one capital letter from [A, B, C, D] 
on its own line.

Format:
<think>
</think>
<LETTER>

Fact: \{fact\}
Question: \{question\}
Options:
A. \{option1\}
B. \{option2\}
C. \{option3\}
D. \{option4\}
\end{alltt}

We use the following prompt to enable reasoning for the same dataset:
\begin{alltt}\small
Answer the following multiple-choice question by 
selecting the single best option (A–D).

After </think>, only output your final decision 
as EXACTLY one capital letter from 
[A, B, C, D] on its own line. Do not output 
any other text after decision.

Format:
<think>
your thinking here
</think>
<LETTER>

Fact: \{fact\}
Question: \{question\}
Options:
A. \{option1\}
B. \{option2\}
C. \{option3\}
D. \{option4\}
\end{alltt}

For the content moderation datasets (Open AI Moderation and XSTest), we use the same prompt for all experiments:
\begin{alltt}\small
You are a safety classification assistant. Given 
an input message, classify it as either:

- "Safe" – if it does not contain harmful, toxic,
or unsafe content.
- "Unsafe" – if it contains hate speech, threats,
self-harm, sexual content, or otherwise unsafe
material.

Respond with only the classification label: 
"Safe" or "Unsafe".

Input:
"\{input\}"
Classification:
\end{alltt}

\section{Additional Experimental Results}

\subsection{Applying Post-Hoc Calibration Methods}\label{sec:post-hoc}

Our approach modifies the model parameters to yield inherently better-calibrated confidence estimates while simultaneously adapting the model to downstream tasks. Post-hoc calibration techniques can then be applied on top of this model to further improve calibration. In this section, we evaluate the combination of our calibration-aware reinforcement learning formulation and standard post-hoc methods. We apply two widely used techniques, Platt Scaling \cite{platt1999probabilistic} and Isotonic Regression \cite{10.1145/775047.775151}, after both vanilla GRPO and our method. Calibration is performed on a subset of OpenAI Moderation dataset, and evaluation is conducted on both the held-out OpenAI samples and XSTest. 

Our results in Table~\ref{tab:post-hoc} show that both post-hoc methods improve the calibration of GRPO-trained models. At the same time, our fine-tuning approach achieves larger and more consistent improvements, and applying post-hoc calibration on top of our method tends to yield the best overall calibration performance. In this sense, post-hoc calibration is orthogonal and complementary to our approach: the combined method offers clear benefits over vanilla GRPO.

\begin{table}[ht]
\centering
\fontsize{10.5}{16}\selectfont
\begin{tabular}{l|l|cc|cc}
\toprule
\multirow{2}{*}{\textbf{}} & \multirow{2}{*}{\textbf{Post-Hoc}} & \multicolumn{2}{c}{\textbf{OpenAI}} & \multicolumn{2}{|c}{\textbf{XSTest}} \\
\cmidrule(lr){3-4} \cmidrule(lr){5-6}
& & GRPO & Ours & GRPO & Ours \\
\midrule
\multirow{3}{*}{\rotatebox{90}{{Qwen-1.7B}}} & None & 12.20 & 8.67 & 20.22 & 15.48 \\
                           & Isotonic & 4.80 & \textbf{3.36} & 12.82 & \textbf{7.24} \\
                           & Platt & 6.22 & 7.60 & 10.89 & 9.69 \\
\midrule
\multirow{3}{*}{\rotatebox{90}{{Qwen-4B}}}   & None & 11.00 & 7.11 & 12.21 & 8.89 \\
                           & Isotonic & 4.85 & \textbf{2.8} & 10.98 & \textbf{3.9} \\
                           & Platt & 7.78 & 6.77 & 4.93 & 5.67 \\
\midrule
\multirow{3}{*}{\rotatebox{90}{{Qwen-8B}}}   & None & 10.68 & 7.11 & 13.54 & 6.41 \\
                           & Isotonic & 4.07 & \textbf{3.78} & \textbf{1.62} & 3.52 \\
                           & Platt & 10.68 & 6.15 & 9.61 & 6.9 \\
\bottomrule
\end{tabular}
\caption{ECE Scores after applying different post-hoc calibration methods to vanilla GRPO and our proposal. }
\label{tab:post-hoc}
\end{table}

\subsection{Additional Calibration Metrics}

For completeness, we report results using two additional calibration metrics: Static Calibration Error (SCE) \cite{Nixon_2019_CVPR_Workshops} and Marginal Calibration Error (MCE) \cite{NEURIPS2019_f8c0c968}. As shown in Table~\ref{tab:additional_metric}, these metrics yield results consistent with the findings reported in the main text. The relative trends across methods are preserved, and the inclusion of SCE and MCE reinforces the conclusions drawn from our combined analysis of distributions, diagrams, and ECE.

\input{tables/add_calib_metric}

\subsection{Complete Plots}

\begin{figure*}[t]
    \centering
    \includegraphics[width=0.89\linewidth]{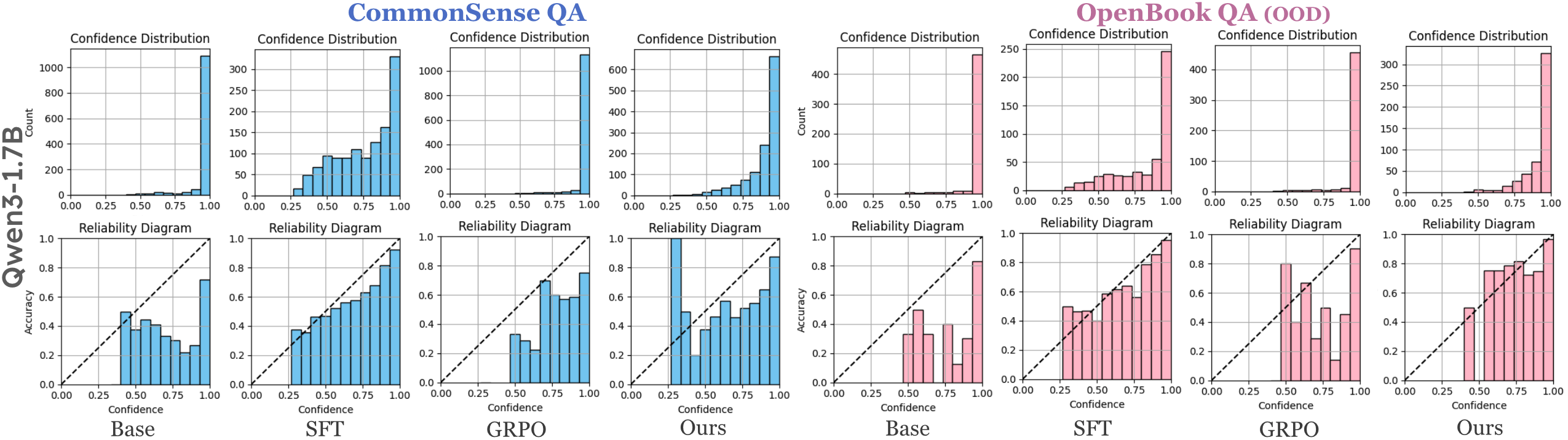}\\[3pt]
    \includegraphics[width=0.89\linewidth]{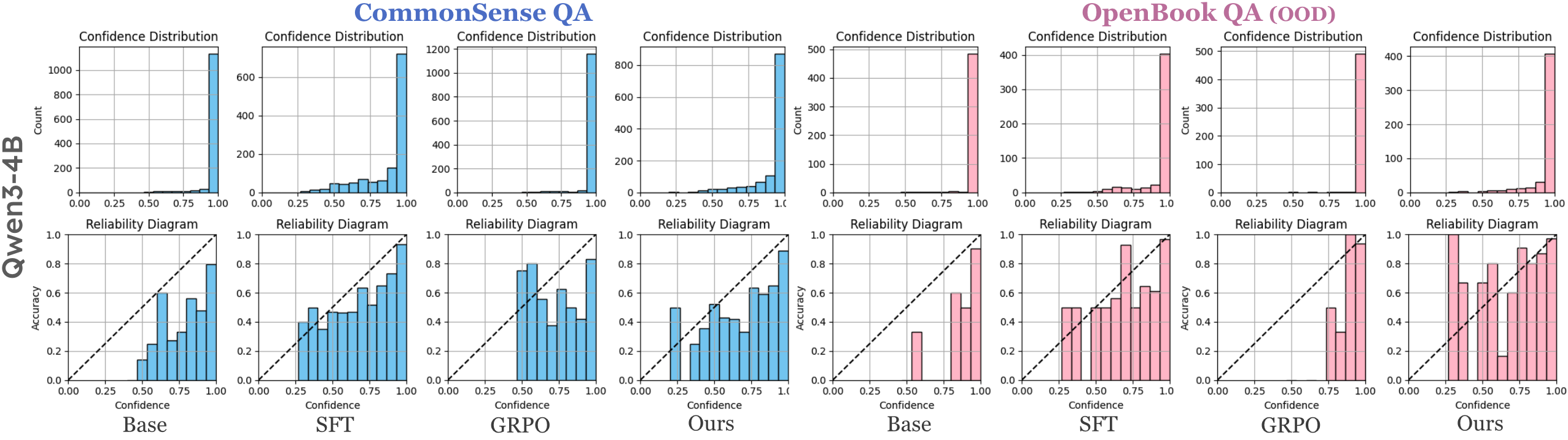}\\[3pt]
    \includegraphics[width=0.89\linewidth]{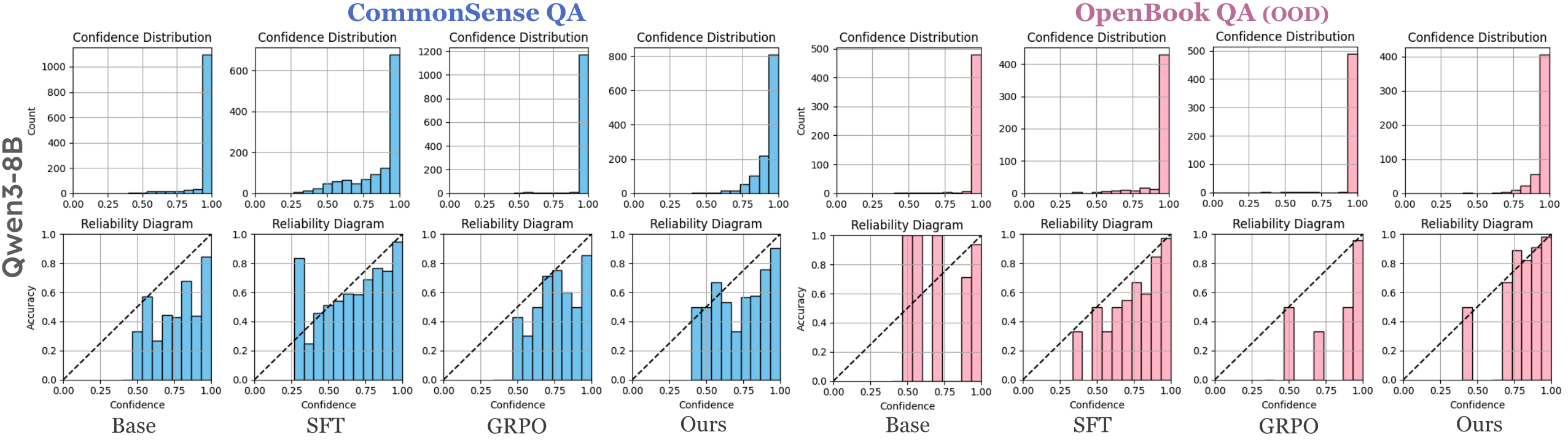}\\[3pt]
    \includegraphics[width=0.89\linewidth]{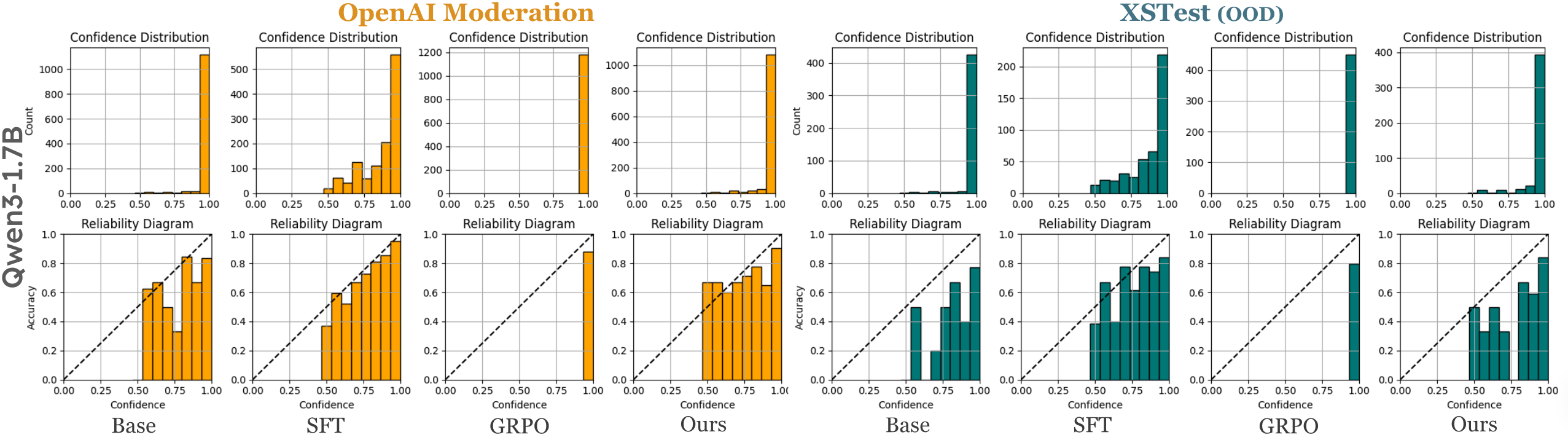}\\[3pt]
    \includegraphics[width=0.89\linewidth]{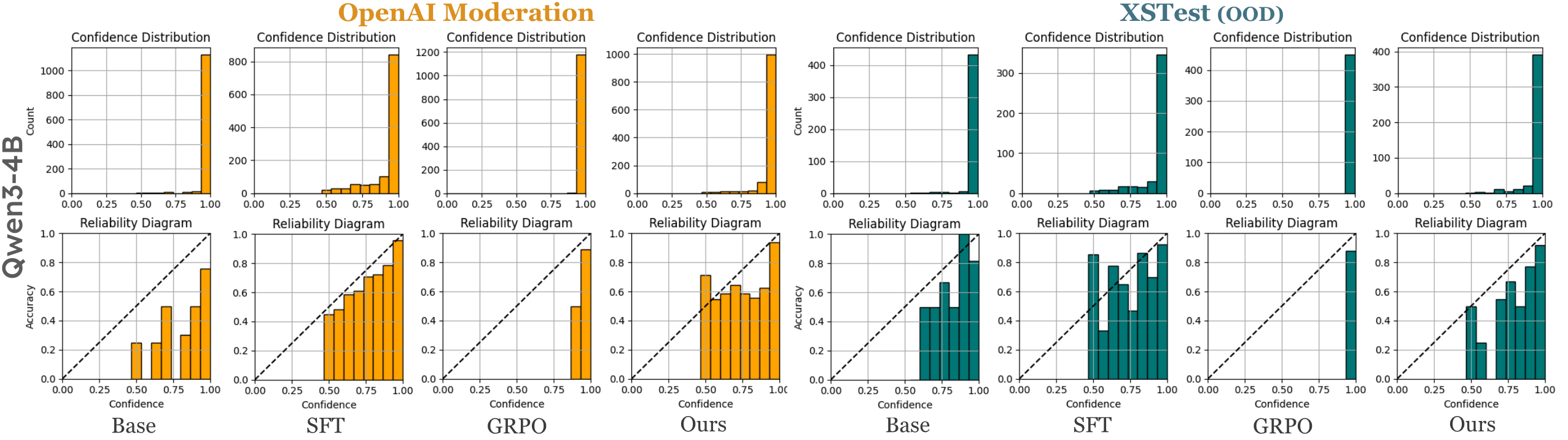}\\[3pt]
    \includegraphics[width=0.89\linewidth]{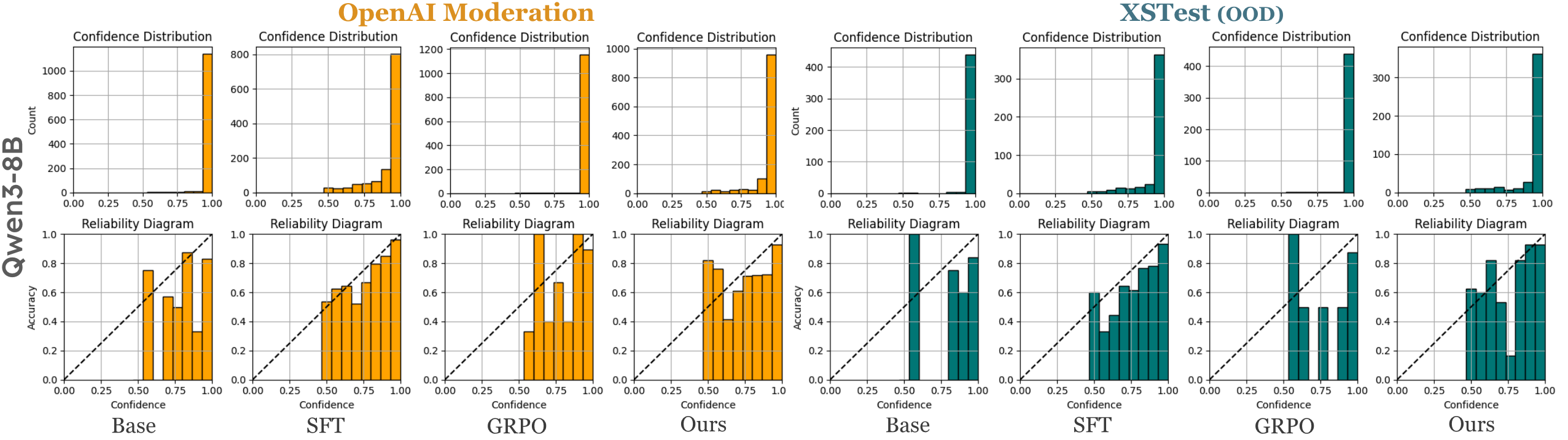}
\vskip -0.1in
    \caption{Reliability diagrams and confidence distributions of all model-dataset pairs.}
    \label{fig:complete_diagrams}
\end{figure*}

We present the complete plots in Figure~\ref{fig:complete_diagrams}.

\section{Additional Considerations} \label{sec:additional_discussion}

\paragraph{Dual-Use of Calibration} Highly calibrated confidence scores provide more accurate estimates of model uncertainty, which can strengthen downstream safety mechanisms. However, the same information could, in principle, be leveraged by adversaries to more efficiently identify decision-boundary regions where the model is most vulnerable to jailbreaks or targeted attacks. This creates a trade-off: while overconfident models pose the risk of high-confidence false negatives, exposing well-calibrated confidence scores may reveal patterns that facilitate targeted attacks.

A practical mitigation is to treat calibrated uncertainty as an internal signal used solely for risk-sensitive routing, escalation, or policy enforcement, rather than exposing it to end users. Although this reduces the likelihood of direct exploitation, it does not fully eliminate the underlying issue. Understanding which aspects of uncertainty information are most susceptible to adversarial use, and developing defenses that preserve the benefits of calibration without enlarging the attack surface, represents an important direction for future work.

\paragraph{Generality Beyond Decision Tasks}
Our method is designed for decision-making tasks in which confidence is associated with a single decision token. This setting covers a broad class of high-impact applications, including safety classification, medical decision support, and many agentic subroutines, where task structure aligns naturally with our formulation.

Extending calibration-aware training to open-ended generation remains an important direction. A straightforward conceptual extension is to adopt a meta-level classification strategy: for a generated long-form answer containing multiple intermediate claims, the model could be prompted to predict the correctness of the entire sequence or of individual claims, and our calibration loss could then be applied to this dedicated correctness token. However, implementing such an approach requires additional infrastructure, such as reliable claim extraction, defining appropriate units of decision, and aggregating confidence across claims, which lies outside the scope of our contribution.

Our work therefore focuses on decision-level calibration rather than structural decomposition of long-form outputs, and we explicitly note this limitation as a promising avenue for future research.

\paragraph{Generalizability Beyond Overconfident Base Models}

Our analysis reveals that base Qwen3 models are overconfident on reasoning-enabled decision making tasks. A follow-up question is whether the standard RLVR algorithm would similarly result in overconfident models when applied to a well-calibrated base model. We discuss this from three perspectives: 

In practice, finding a modern LLM that is both instruction-tuned and well-calibrated is difficult \cite{xiao2025restoring, huang2025calibrated, leng2025taming}. Since this overconfidence pattern is common across widely used foundation models, we expect our analysis and conclusions to apply broadly.

Even if a well-calibrated base model were available, there are structural reasons why RLVR would remain overconfident. Our analysis shows that the decision token primarily acts as an extraction step (Section 4): it deterministically reflects the conclusion of the reasoning trace rather than expressing epistemic uncertainty. This behaviour is inherently a property of task formulation rather than the base model. Thus, the overconfidence observed under RLVR is a structural consequence of RL on reasoning-enabled decision-making tasks, and is not tied to the initial calibration state of the model.  

Lastly, even a perfectly calibrated decision token would be pushed toward overconfidence under vanilla RLVR. For instance, if the decision token probabilities after reasoning were initially well-calibrated (e.g., outputting the correct label with probability 0.7), the standard RLVR objective would still increase that probability whenever the trace receives a positive reward. This repeated reinforcement naturally pushes correct decisions toward the overconfidence regime, thereby hurting calibration. Thus, vanilla RLVR is inherently biased toward overconfidence on correct trajectories, regardless of the starting model.

For these reasons, we expect the overconfidence behavior of RLVR to generalize beyond the specific overconfident base models used in our experiments. We leave experimental validation of this as future work due to legal and computational constraints.

%% file: sections/6-RelatedWorks.tex
\section{Related Works}

\paragraph{Confidence Estimation in Generative Models.}
Confidence estimation for generative models has been extensively studied, with a wide range of methods proposed to quantify and calibrate model confidence. The goal is to design an estimator whose confidence scores accurately reflect prediction correctness \cite{bakman2025uncertaintyfeaturegapsepistemic}. 
Existing approaches can be broadly grouped into three categories. Logit-based methods compute confidence directly from token log-probabilities and aggregate them across the generation \cite{bakman2024mars, yaldiz2024designlearntrainablescoring, malinin2021uncertainty, tokensar}.  
Sampling-based methods assess the consistency among multiple sampled generations, where higher agreement implies higher confidence \cite{kuhn2023semantic, lin2023generating, nikitin2024kernel}.  
Verbalized confidence methods prompt the model to explicitly report its self-assessed confidence \cite{kadavath2022language, tian-etal-2023-just}.  
Among these, logit-based methods are the most computationally efficient, as they avoid repeated sampling or secondary prompting, making them particularly practical for large-scale decision-making applications.

In this work, our aim is not to design a new uncertainty estimation algorithm, but to improve model calibration under an existing, efficient confidence definition: the probability of the final decision token as the model’s confidence. We adopt this approach because its superior computational efficiency and single-pass latency are highly desirable properties for real-world deployment, making it arguably the most practically valuable method for using LLMs into low-latency decision systems.

\paragraph{Calibration-Aware Fine-Tuning.}
Several recent studies have explored fine-tuning approaches that explicitly optimize for calibrated confidence. Most of these works adopt verbalized confidence as the estimation mechanism and adjust training to align self-reported confidence with empirical correctness. For instance, \citet{kadavath2022language, lin2022teaching, liu-etal-2024-llms-learn-uncertainty} apply supervised fine-tuning to verbalized confidence data, while \citet{stangel2025rewardingdoubtreinforcementlearning, tao-etal-2024-trust, xu-etal-2024-sayself, damani2025binaryrewardstraininglms, NEURIPS2024_4b8eaf3b} explore reinforcement learning to calibrate confidence expression. Although these approaches share the goal of improving calibration, they are fundamentally distinct from our work, as they operate on a different confidence modality. Our focus lies on logit-based confidence calibration, which remains underexplored despite its computational advantages and prevalence in real-world deployment scenarios.

A recent work \cite{xiao2025restoring} investigates probability-based calibration rather than verbalized confidence. They study calibration in aligned LLMs and show that preference-alignment objectives (e.g., RLHF or DPO) tend to degrade calibration. They propose a calibration-aware fine-tuning method to restore calibration by modifying the supervised fine-tuning loss. However, their setup does not involve reasoning traces and remains SFT-based, making their approach conceptually similar to our SFT baseline. In contrast, our method intervenes directly within reinforcement learning: we modify GRPO itself to calibrate the probability of the final decision token while preserving reasoning–decision consistency.

\paragraph{Post-Hoc Calibration.}
Beyond fine-tuning and uncertainty estimation, a large body of work from the classification literature focuses on post-hoc calibration \cite{10.5555/3305381.3305518}: adjusting model output probabilities after training to improve alignment with empirical accuracy. Classical techniques include temperature scaling \cite{10.5555/3305381.3305518}, isotonic regression \cite{10.1145/775047.775151}, and Platt scaling \cite{platt1999probabilistic}. \citet{zhou2024batch} proposes Batch Calibration for in-context learning scenarios. Recent work \cite{liu2025on} extends these ideas to LLM-based guard models such as Llama Guard \cite{inan2023llamaguardllmbasedinputoutput} and WildGuard \cite{NEURIPS2024_0f69b4b9}, evaluating the impact of post-hoc methods on safety and moderation tasks. 
Another work \cite{wang2024calibratingverbalizedprobabilitieslarge} applies post-hoc calibration directly to LLM-generated verbalized probabilities using the inverted softmax trick: model probabilities are inverted to recover logits, after which temperature scaling is applied to produce calibrated confidence estimates.
Our work differs substantially in focus: we study calibration during the fine-tuning process itself, rather than adjusting already fine-tuned models. Importantly, our approach is orthogonal to post-hoc methods. Such calibration techniques can be applied on top of our fine-tuning procedure. In Section \ref{sec:post-hoc}, we provide experimental results showing that applying post-hoc calibration on top of our method tends to yield the best overall calibration performance.

%% file: tables/add_calib_metric.tex
\begin{table}[!htbp]
\centering
\fontsize{7.5}{10.5}\selectfont
\begin{tabular}{l|l| cc |cc |cc}
\toprule
  & &   \multicolumn{2}{c|}{\textbf{ Qwen3-1.7B}}  & \multicolumn{2}{c|}{\textbf{Qwen3-4B}} &   \multicolumn{2}{c}{\textbf{ Qwen3-8B}} \\
   & & SCE & MCE & SCE & MCE & SCE & MCE  \\
   & & ($\downarrow$) & ($\downarrow$)& ($\downarrow$) & ($\downarrow$) & ($\downarrow$) & ($\downarrow$)  \\
\midrule
\multirow{4}{*}{\rotatebox{90}{\textbf{CSQA$^\dagger$}}}
&	Base 	& 11.56	& 16.68	& 7.82	& 12.23	& 6.42	& 11.01	\\
&	SFT 	& 3.57	& 5.53	& 3.75	& 6.05	& 2.41	& 3.97	\\
&	GRPO 	& 9.11	& 13.66	& 6.50	& 10.75	& 5.60	& 9.57	\\
&	Ours 	& 6.20	& 9.21	& 5.03	& 8.41	& 4.07	& 6.34	\\
\midrule
\multirow{4}{*}{\rotatebox{90}{\textbf{OBQA$^*$}}}
&	Base 	& 8.75	& 14.16	& 4.49	& 8.29	& 2.97	& 6.89	\\
&	SFT 	& 4.72	& 8.17	& 2.73	& 5.24	& 2.06	& 5.06	\\
&	GRPO 	& 5.23	& 8.92	& 2.94	& 6.52	& 2.02	& 4.63	\\
&	Ours 	& 3.15	& 5.16	& 1.67	& 2.71	& 1.60	& 2.45	\\
\bottomrule
\end{tabular}
\caption{Calibration metrics (SCE and MCE, values in \%) for base, SFT, GRPO, and our proposal. While $\dagger$ indicates in-domain, * denotes out-of-domain datasets.}
\label{tab:additional_metric}
\end{table}